\pdfoutput=1

\PassOptionsToPackage{dvipsnames}{xcolor}

\documentclass[11pt]{article}

\usepackage[final]{acl}
\usepackage{array, float, multirow}
\usepackage{times}
\usepackage{latexsym}

\usepackage{amsmath}
\usepackage{amssymb}
\usepackage{amsfonts}
\usepackage{subfig}  

\usepackage{hyperref}
\hypersetup{
    colorlinks=true,
    linkcolor=blue,
    filecolor=magenta,      
    urlcolor=blue,
    pdftitle={Overleaf Example},
    pdfpagemode=FullScreen,
    }

\usepackage{graphicx}
\usepackage[T1]{fontenc}

\usepackage[utf8]{inputenc}

\usepackage{microtype}

\usepackage{inconsolata}

\title{UnibucLLM: Harnessing LLMs for Automated Prediction of Item Difficulty and Response Time for Multiple-Choice Questions}

\author{Ana-Cristina Rogoz \and Radu Tudor Ionescu\\
Department of Computer Science\\
University of Bucharest\\
14 Academiei, Bucharest, Romania\\
\texttt{raducu.ionescu@gmail.com} \\}

\begin{document}
\maketitle
\begin{abstract}
This work explores a novel data augmentation method based on Large Language Models (LLMs) for predicting item difficulty and response time of retired USMLE Multiple-Choice Questions (MCQs) in the BEA 2024 Shared Task. Our approach is based on augmenting the dataset with answers from zero-shot LLMs (Falcon, Meditron, Mistral) and employing transformer-based models based on six alternative feature combinations. The results suggest that predicting the difficulty of questions is more challenging. Notably, our top performing methods consistently include the question text, and benefit from the variability of LLM answers, highlighting the potential of LLMs for improving automated assessment in medical licensing exams. We make our code available at: \url{https://github.com/ana-rogoz/BEA-2024}.
\end{abstract}

\section{Introduction}

High-stakes medical licensing exams, like the United States Medical Licensing Examination (USMLE), require well-crafted questions to accurately assess a candidate's knowledge and skills. Traditionally, determining item difficulty and response time (average time to answer) relied on pretesting, which can be carried out by embedding new items alongside scored items in live exams. However, this method has been recognized as impractical due to resource limitations \cite{settles-etal-2020-machine}.

This year's Workshop on Innovative Use of NLP for Building Educational Applications (BEA 2024) directly addresses this problem and its resource limitations by proposing a shared task on Automated Prediction of Item Difficulty and Item Response Time for USMLE exam items. This initiative fosters collaboration and innovation in developing reliable prediction methods, while also contributing to creating more efficient, secure, and informative medical licensing exams.

This paper details our participation in the shared task \cite{yaneva-etal-2024-DART-MCQ}, where we investigated the use of Large Language Models (LLMs) to predict difficulty and response time for retired USMLE Multiple-Choice Questions (MCQs). Our main contribution is to augment the dataset by incorporating answer choices generated by several zero-shot LLMs (Falcon, Meditron, Mistral). To solve the two prediction tasks (question response time prediction, question difficulty prediction), we employ transformer-based models that alternatively employ six different feature combinations. Our findings indicate that predicting question difficulty proves to be a more complex task. Interestingly, the most successful models consistently incorporate the question text, and benefit from the augmentation based on LLM-generated answers. Our results highlight the potential of LLMs to enhance automated assessment methods in medical licensing exams.

We also present post-competition methods that obtain better results than the originally submitted models. These newer models are aimed at addressing overfitting and our wrong choice of features.

\section{Related work}

The need for alternatives to the traditional processes motivates the exploration of new methods for estimating item difficulty and response time. Recent research \cite{ha-etal-2019-predicting, yaneva-etal-2020-predicting, xue-etal-2020-predicting, Baldwin-JEM-2021, yaneva-etal-2021-using} suggests promising results using machine learning models trained on item text data to predict these characteristics. 

\begin{figure*}[t!]
     \centering
    \includegraphics[width=1.0\textwidth]{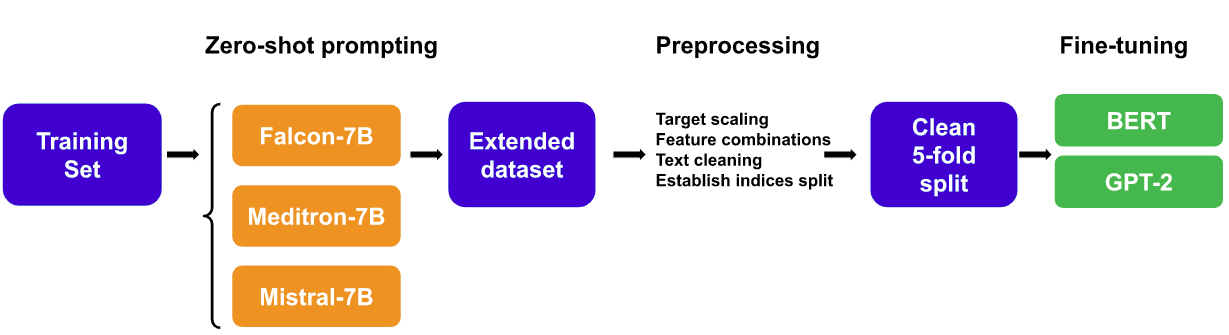}
    \caption{An overview of the data preprocessing and model training workflow for predicting item difficulty and response time of medical exam questions. The initial dataset is enriched with zero-shot prompted responses generated by Large Language Models (LLMs). We then perform preprocessing over the augmented dataset by scaling the target labels, adding new feature combinations, text cleaning and establishing the split for cross-validation. Finally, two alternative transformer-based models are fine-tuned on the augmented data.}
    \label{fig:overview}
\end{figure*}

One of the seminal studies in this direction \cite{ha-etal-2019-predicting} investigated the feasibility of using machine learning models to predict both item difficulty and response time for multiple-choice questions in a high-stakes medical exam. The authors focused on extracting various features from the question text data, including linguistic features and embedding types. Their models were then trained to predict these difficulty and response time characteristics. The encouraging results from this study suggest that machine learning offers a promising alternative to traditional, resource-intensive pretesting methods for estimating these important exam design elements. 

While the prior studies focused on predicting difficulty and response time separately, \citet{xue-etal-2020-predicting} explored a method that could predict both simultaneously, using transfer learning. Their research suggests that this approach offers potential benefits in terms of efficiency.

In addition to predicting difficulty and response time, researchers explored another valuable application of machine learning: item survival prediction \cite{yaneva-etal-2020-predicting}. This task focuses on estimating the likelihood of an item to be included in the final exam based on its difficulty and other question characteristics, and highlights the versatility of machine learning for various stages of exam design. 

Another approach was presented by \citet{Baldwin-JEM-2021}, who study the use of linguistic features to predict the response process complexity, which refers to the mental steps required to answer a medical MCQ. 

Instead of predicting difficulty, \citet{yaneva-etal-2021-using} leveraged the use of linguistic features to predict the response process complexity associated with answering medical MCQs. Their work sheds light on the underlying factors that contribute to the difficulty of these questions.

In summary, automated approaches offer several advantages, such as efficiency (predicting from text eliminates the need for pretesting, saving time and resources), security (reduced reliance on pretesting minimizes the risk of question exposure), and scalability (automated methods allow for creating larger pools of high-quality questions). Therefore, continuously validating the use of machine learning to replace traditional methods is currently an active research topic. 

\section{Methods}

We start by annotating the original dataset with answers obtained by prompting LLMs in a zero-shot setup. The extended dataset is further processed by scaling the target labels, creating additional feature combinations, text cleaning, and setting the data split for cross-validation. The cleaned dataset is employed to fine-tune two transformer-based models. The end-to-end overview of the employed framework is presented in Figure~\ref{fig:overview}. Below, we describe each step of our pipeline in detail.

\begin{table*}
\setlength\tabcolsep{2.5pt}
\centering
 \begin{tabular}{|p{0.06\linewidth}|p{0.295\linewidth}|p{0.295\linewidth}|p{0.295\linewidth}|}
 \hline
  \#Item & Falcon & Meditron & Mistral \\ 
 \hline\hline
  391 & The correct answer is: C. Weight loss program.
The correct answer is: C. & 
The correct answer is option A. The patient has a history of hy & 
The correct answer is D. Antihypertensive therapy.
The patient has \\
\hline
148 & The answer is:
A. Common fibular (peroneal),
The common & 
The correct answer is option A. The common fibular nerve is & 
The correct answer is A. Common fibular (peroneal).

The common \\
\hline
562 & The answer is:
A. A, B. B, C. C, D. & 
The correct answer is option D. The correct answer is option D. & 

The correct answer is D. D.

The patient has a\\
 \hline
 \end{tabular}
 \caption{Examples of Falcon, Meditron and Mistral answers, when prompted with USMLE questions together with the multiple-choice answers. The examples are not truncated (although it often seems so).}
\label{LLMs-answers}
\end{table*}

\begin{table}[!hbt]
\setlength\tabcolsep{2.5pt}
\centering
\begin{tabular}{|p{0.35\linewidth}|p{0.6\linewidth}|}
     \hline
     Feature name & Description \\
     \hline \hline
     ItemNum &  Index\\
     \hline
    ItemText & Question text\\  
    \hline
    Answer\_[A-J] & Multiple choice answers\\
    \hline
    Answer\_Key & Single value between A-J\\ 
    \hline
    Answer\_Text & The text of the correct answer\\ 
    \hline
    ItemType & Text or PIX (i.e. image) \\
    \hline
    EXAM & Step\_[1, 2, 3] \\
    \hline
    Difficulty & Real value indicating question difficulty. \\
    \hline
    Response\_Time & Integer value indicating  mean response time (s).\\
    \hline
\end{tabular}
 \caption{Initial set of features from the original shared-task dataset.}
\label{tab_original_features}
\end{table}

\subsection{Zero-Shot Prompting} 

We conjecture that LLMs can be employed to provide answers to the questions that need to be evaluated in terms of difficulty and response time, and the returned answers can be harnessed to better solve the prediction tasks. For instance, the number of LLMs that give correct answers to a question can be a strong indicator for the difficulty level of the respective question. To this end, we rely on three LLMs to obtain the answers, namely Falcon-7B \cite{falcon40b}, Meditron-7B \cite{chen2023meditron70b} and Mistral-7B \cite{jiang2023mistral}. We resort to the use of models with 7B parameters, due to our computing resource limitations. However, we compensate for the use of lighter LLMs by integrating multiple models. While Meditron-7B is specialized on the medical domain, which perfectly suits the provided shared task data, the other LLMs are general purpose models. These choices are aimed at enhancing the \emph{diversity} of the models, which was previously reported as a relevant aspect when constructing ensembles \cite{georgescu2023diversity}. Hence, by combining the outputs of the three LLMs, we aim to leverage the complementary strengths of all models. The selected LLMs are trained on distinct datasets, and they exhibit different capabilities in reasoning, factual recall, or creative text generation. By employing diverse models, we aim to reduce the influence of biases learned by individual models, thus achieving a higher generalization capacity. For each sample, we prompt the three LLMs in the following manner: 
\begin{verbatim}
PROMPT: "You are a student taking the 
USMLE exam. Your task is to answer the 
following question with one of the 
multiple choices. 

$ItemStem_Text

A.$Answer_A, 
B.$Answer_B, 
..."
\end{verbatim}

Building on the provided prompts, Table~\ref{LLMs-answers} showcases example responses from the three LLMs (Falcon-7B, Meditron-7B, and Mistral-7B). Interestingly, we observe a wide spectrum of agreement, ranging from all three models providing identical answers to complete divergence in their responses.

\subsection{Preprocessing}

To ensure consistent scaling across labels, we normalize the ``Response\_Time'' and ``Difficulty'' target labels to a common range between 0 and 1. Following the scaling of target variables, we apply additional preprocessing steps to the LLM outputs. To improve performance and data consistency, we cleaned the LLM answer texts by removing any extra spaces and new line characters. 

To ensure the reproducibility of our results, we provide the preprocessed and augmented dataset, containing both training and test sets, at \url{https://github.com/ana-rogoz/BEA-2024}. 

\begin{figure*}[hbt!]
    \centering
    \begin{minipage}{0.5\textwidth}
        \centering
        \includegraphics[width=\textwidth]{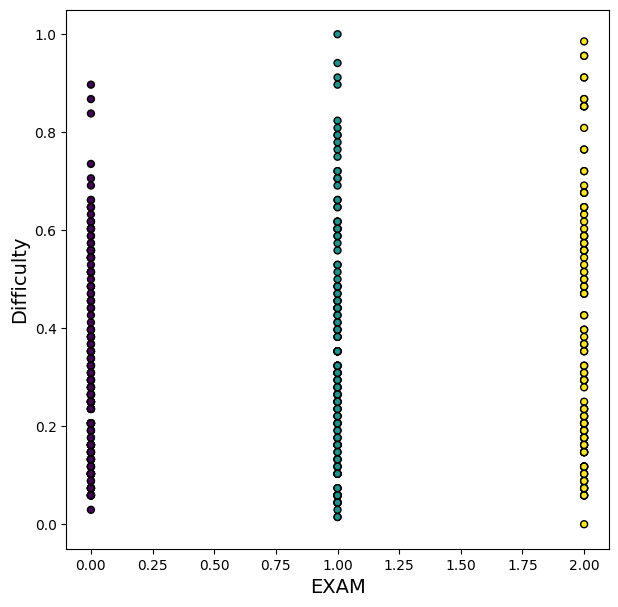} 
    \end{minipage}\hfill
    \begin{minipage}{0.5\textwidth}
        \centering
        \includegraphics[width=\textwidth]{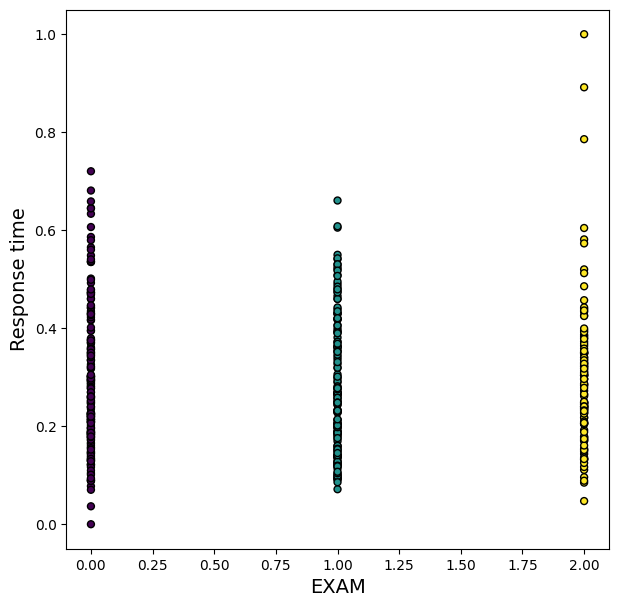} 
    \end{minipage}
    \caption{\textbf{Left}: Correlation between the EXAM integer feature and the difficulty label. \textbf{Right}: Correlation between the EXAM integer feature and the response time label.}
    \label{fig:EXAM-correlation}
\end{figure*}

\begin{figure*}[hbt!]
    \centering
    \begin{minipage}{0.5\textwidth}
        \centering
        \includegraphics[width=\textwidth]{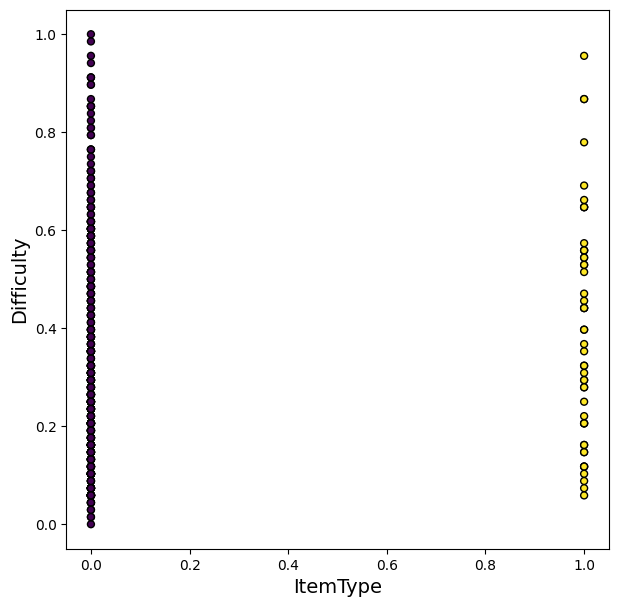} 
    \end{minipage}\hfill
    \begin{minipage}{0.5\textwidth}
        \centering
        \includegraphics[width=\textwidth]{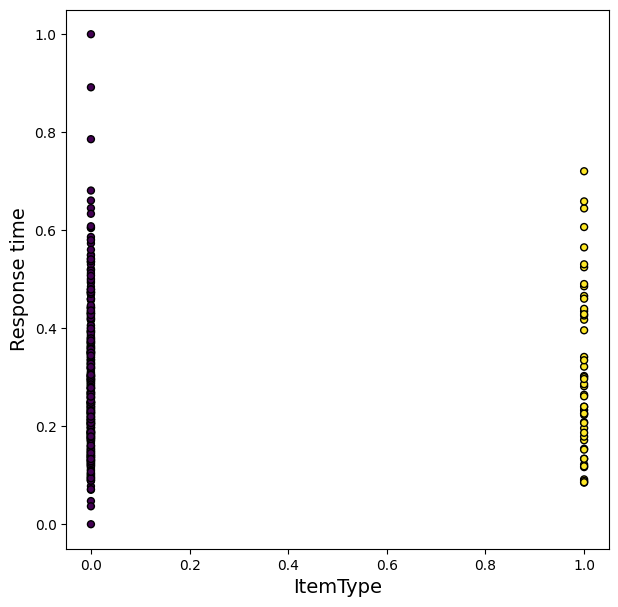} 
    \end{minipage}
    \caption{\textbf{Left}: Correlation between the ItemType integer feature and the difficulty label. \textbf{Right}: Correlation between the ItemType integer feature and the response time label.}
    \label{fig:item-correlation}
\end{figure*}

\begin{figure*} [hbt!]
    \centering
    \begin{minipage}{0.5\textwidth}
        \centering
        \includegraphics[width=\textwidth]{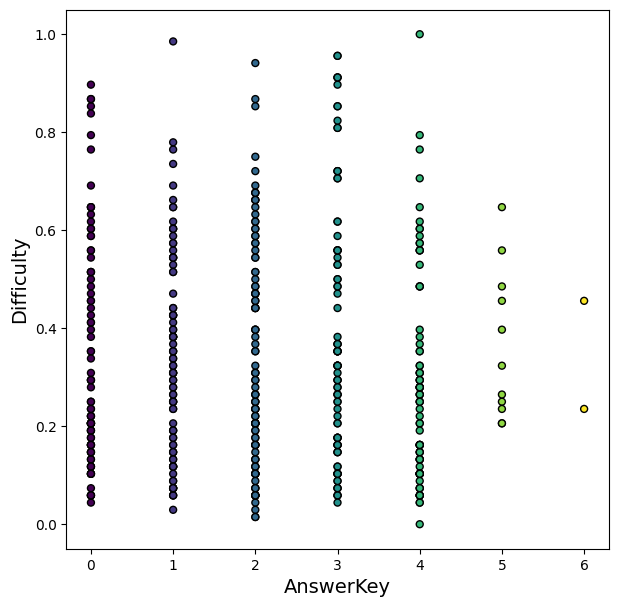} 
    \end{minipage}\hfill
    \begin{minipage}{0.5\textwidth}
        \centering
        \includegraphics[width=\textwidth]{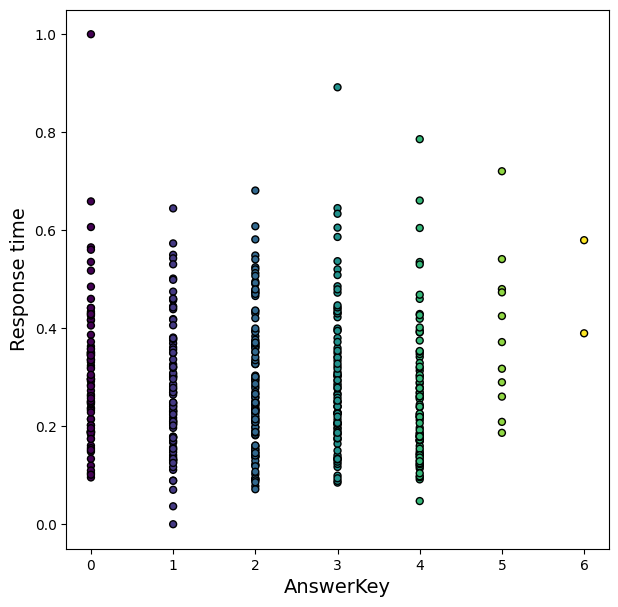} 
    \end{minipage}
    \caption{\textbf{Left}: Correlation between the AnswerKey integer feature and the difficulty label. \textbf{Right}: Correlation between the AnswerKey integer feature and the response time label.}
    \label{fig:answer-correlation}
\end{figure*}

\subsection{Data Engineering} 

A detailed breakdown of the available features can be found in Table~\ref{tab_original_features}. We checked how well input features correlate to the target Response\_Time and Difficulty values, and concluded that the EXAM, AnswerKey and ItemType columns display no correlation, as shown in Figures \ref{fig:EXAM-correlation}, \ref{fig:item-correlation} and \ref{fig:answer-correlation}, respectively. Thus, before the competition, we decided to exclude these columns from all our experiments, except for one baseline that includes all original features. However, this decision overlooks an important insight: although the AnswerKey alone does not correlate with the labels, it could represent a very useful feature when combined with LLM answers. This is because an LLM can answer ``The correct answer is D.'', and comparing this answer with the AnswerKey feature can tell us if the LLM was able to correctly identify the correct answer or not. To this end, we combine the AnswerKey feature with LLM features in our post-competition models.

To enrich the input data provided to our models, we engineer seven new features, as presented in Table~\ref{tab_new_features}. These features combine original dataset features with the LLM-generated answers. This process aims to capture a more comprehensive representation of the problem for the trained models.

\begin{table}[!t]
\setlength\tabcolsep{3.0pt}
\centering
\begin{tabular}{|p{0.3\linewidth}|p{0.6\linewidth}|}
     \hline
     Feature set & Merged features \\
     \hline \hline
     all & All initial feature columns \\
     \hline
    q\_answers & ItemText, Answer\_{*}, Answer\_Text\\  
    \hline
    answers & Answer\_{*}\\
    \hline
    q\_a & ItemText, AnswerText\\ 
    \hline
    llms\_a & LLM answers, AnswerText\\ 
    \hline
    q\_llms\_a & ItemText, LLM answers, AnswerText \\
    \hline
    q\_llms\_a\_key$^\diamond$ & ItemText, LLM answers, AnswerText, AnswerKey \\
    \hline
\end{tabular}
 \caption{Combinations of features that are alternatively used to train our models. The $\diamond$ symbol indicates the feature set is added post-competition.}
\label{tab_new_features}
\end{table}

To mitigate the limitations of the very small dataset size, we employ a 5-fold cross-validation procedure to train our models. This technique involves shuffling the data and splitting it into five fixed equally-sized subsets.
Each fold is then used for training and validation in turn, providing a robust evaluation of the models. 
The final extended and shuffled dataset is part of our publicly available repository. 

\subsection{Models}

Our work focuses on training and applying different models to the two regression tasks, namely predicting response time and difficulty of medical questions. We utilize two transformer-based approaches, well-suited for learning complex relationships. The models are trained on the new sets of constructed features, which are detailed in Table~\ref{tab_new_features}. We also include a basic linear modeling approach as baseline. After the competition, we decided to employ a model that uses frozen transformer-based features and trains only a linear model on top of the deep features. This decision is aimed at addressing the potential of overfitting transformer-based models to the very small dataset available for the competition.

\noindent
\textbf{Fine-tuned BERT.} Our first method employs a fine-tuned Bidirectional Encoder Representations from Transformers (BERT) model \cite{DBLP:journals/corr/abs-1810-04805} for the regression tasks. We leverage the pre-trained BERT encoder to generate contextualized representations for each text input as 768-dimensional vectors. 
However, instead of the standard classification head, we implement a single-neuron regression head. Finally, a sigmoid activation function is applied to the output layer, ensuring the predictions fall within the desired range of 0 to 1.

\noindent
\textbf{Fine-tuned GPT-2.} 
Similar to the BERT-based approach, our second method fine-tunes a GPT-2 model \cite{radford2019language} for regression. We utilize the corresponding GPT-2 tokenizer to convert text inputs into numerical representations. The pre-trained GPT-2 model undergoes further training (fine-tuning) with a single-neuron output layer at the end. Once again, we employ a sigmoid activation function to ensure the model's predictions fall within the interval $[0, 1]$.

\noindent
\textbf{$\boldsymbol{\nu}$-Support Vector Regression + TF-IDF.} In addition to transformer-based approaches, we investigate a linear regression method, namely $\nu$-Support Vector Regression ($\nu$-SVR) \cite{Scholkopf-NC-2000}. We experiment with two shallow feature extraction techniques, namely TF-IDF and TF-IDF combined with Principal Component Analysis (PCA), focusing on the statistical properties of words in a document. 

\noindent
\textbf{$\boldsymbol{\nu}$-Support Vector Regression + BERT.}
Fine-tuning large models, e.g.~BERT or GPT-2, on small datasets is prone to overfitting. To mitigate overfitting, an alternative to end-to-end fine-tuning is keeping the pre-trained layers frozen, and training only the last regression layer. To this end, we propose a model that employs BERT-based embeddings and trains a $\nu$-SVR model on top, an approach that is also known as \emph{linear probing}. As input to the BERT model, we consider LLM answers with and without the AnswerKey feature. The resulting $\nu$-SVR+BERT models are added post-competition. 

\section{Experiments}

\subsection{Dataset}
In the BEA 2024 Shared Task, the dataset provided by the organizers consists of retired Multiple-Choice Questions (MCQs) from the United States Medical Licensing Examination. The data is divided into two distinct subsets: an initial training set of 466 samples and a separate test set of 201 samples, which is used to evaluate the participants.

\subsection{Evaluation}

We assess the performance levels of our methods using two complementary metrics: the mean squared error (MSE) and the Kendall $\tau$ correlation. MSE measures the average squared difference between predicted and actual values, indicating how well the model fits the data, while the Kendall $\tau$ correlation evaluates the model's ability to capture the general trend of the data, providing insights into its generalization capability.

\subsection{Hyperparameter Tuning}
\label{sec:hyperparams}
The hyperparameters of all models are determined via grid search.
For the transformer-based methods (BERT, GPT-2), we employ a grid search over the maximum number of input tokens in $\{100, 150, 200, 250, 300, 350, 400, 512\}$, learning rate values in $\{10^{-4}, 5\cdot 10^{-4}, 10^{-5}, 5\cdot 10^{-5}, 10^{-6}, 5\cdot 10^{-6}\}$, and number of training epochs in $\{5, 10, 15, 20\}$. The models are optimized using the AdamW optimizer \cite{Loshchilov-ICLR-2019} on mini-batches of 32 samples. For the $\nu$-SVR approaches, we employ a grid search over the parameter $C$ in the set $\{0.01, 0.1, 0.5, 1, 5, 10, 50, 100\}$ and values of $\nu$ in $\{0.1, 0.2, 0.3, 0.4, 0.5\}$. The complete hyperparameter setup for our experiments, as well as the methods themselves, are available as part of our publicly available repository: \url{https://github.com/ana-rogoz/BEA-2024}.

\begin{table*}[!t]
\setlength\tabcolsep{3.0pt}
\begin{center}
\begin{tabular}{|l|l|l|c|c|c|}
\hline
Task & Model & Features &  MSE $\downarrow$ & Kendall $\tau$ $\uparrow$ & Run \\
\hline 
\hline
\multirow{16}{*}{Response Time} & \multirow{6}{*}{{BERT}} & all & $0.0151\pm0.0016$	& $0.3810\pm0.0543$ & \\ 
\cline{3-6}
& & {\color{red}\textbf{q\_answers}} & ${0.0148}\pm0.0011$	& {\color{red}$\mathbf{0.4232}\pm0.0350$} & 1\\
\cline{3-6}
& & answers & $0.0190\pm0.0017$& $0.1334\pm0.0241$ & \\
\cline{3-6}
& & q\_a & $0.0149\pm0.0010$ & $0.3718\pm0.0452$	& \\
\cline{3-6}
& & llms\_a & $0.0171\pm0.0003$  &	$0.2401\pm0.0467$ & \\
\cline{3-6}
& & q\_llms\_a & $0.0150\pm0.0012$
	& ${0.3912}\pm0.0414$ & 2 \\
\cline{2-6}
& \multirow{6}{*}{{GPT-2}} & all & $0.0245\pm0.0085$	& $0.3550\pm0.0876$ & \\
\cline{3-6}
& & q\_answers & $0.0157\pm0.0023$ &	${0.4029}\pm0.0458$ & 3\\
\cline{3-6}
& & answers & $0.0231\pm0.0041$ &	$0.0703\pm0.0404$ & \\
\cline{3-6}
& & q\_a & $0.0238\pm0.0049$	& $0.2949\pm0.0766$ & \\
\cline{3-6}
& & llms\_a &  $0.0292\pm0.0102$&	$0.1417\pm0.0497$ & \\
\cline{3-6}
& & q\_llms\_a & $0.0249\pm0.0044$ &	$0.2536\pm0.0984$ & \\
\cline{2-6}
& \multirow{4}{*}{SVR} & {\color{RoyalBlue}{q\_llms\_a + BERT}} & $0.0134\pm0.0011$ & {\color{RoyalBlue}$0.4127\pm0.0362$} & $*$ \\
\cline{3-6}
& & {\color{ForestGreen}q\_llms\_a\_key + BERT} & $\mathbf{0.0132}\pm0.0012$ & {\color{ForestGreen}$0.4141\pm0.0289$} & $*$ \\
\cline{3-6}
& & {q\_a + TF-IDF} & $0.0254\pm0.0017$   & $0.1532\pm0.0241$ &\\ 
\cline{3-6}
& & {q\_a + TF-IDF + PCA} &  $0.0294\pm0.0017$ & $0.1132\pm0.0652$ & \\
\hline
\hline
\multirow{16}{*}{Difficulty} & \multirow{6}{*}{{BERT}} & all\_input & $0.0534\pm0.0101$ & $0.0780\pm0.0469$ & \\ 
\cline{3-6}
& & q\_answers & $0.0534\pm0.0102$	& $0.0570\pm0.0862$ & \\
\cline{3-6}
& & answers & $0.0522\pm0.0107$ &	$0.0795\pm0.0481$ & \\ 
\cline{3-6}
& & q\_a & $0.0538\pm0.0092$& $0.0812\pm0.0189$ & \\ 
\cline{3-6}
& & llms\_a & $0.0562\pm0.0105$ 	&  $0.0204\pm0.0610$ & \\
\cline{3-6}
& & {\color{ForestGreen}q\_llms\_a} & $\mathbf{0.0500}\pm0.0093$ & {\color{ForestGreen}${0.1470}\pm0.0447$} & 1 \\
\cline{2-6}
& \multirow{6}{*}{GPT-2} & all\_input & $0.0700\pm0.0080$ & $0.0727\pm0.0640$ & \\
\cline{3-6}
& & q\_answers & $0.0659\pm0.0052$ & $0.1155\pm0.0208$ & 2\\
\cline{3-6}
& & answers & $0.0571\pm0.0130$& $0.0323\pm0.0518$ & \\
\cline{3-6}
& & q\_a & $0.0623\pm0.0059$& $0.0802\pm0.0507$ & \\
\cline{3-6}
& & llms\_a & $0.0707\pm0.0377$& $0.1129\pm0.0472$ & \\
\cline{3-6}
& & {\color{RoyalBlue}q\_llms\_a} & $0.0599\pm0.0142$ & {\color{RoyalBlue}$0.1259\pm0.0333$} & 3\\
\cline{2-6}
& \multirow{4}{*}{SVR} & {q\_llms\_a + BERT} & $0.0576\pm0.0087$& $0.1102\pm0.0665$& $*$\\
\cline{3-6}
& & {\color{red}\textbf{q\_llms\_a\_key + BERT}} & $0.0534\pm0.0067$& {\color{red}$\mathbf{0.1592}\pm0.0616$}& $*$\\
\cline{3-6}
& & {q\_a + TF-IDF} &  $0.0551\pm0.0033$ & $-0.0895\pm0.0305$ & \\ 
\cline{3-6}
& & {q\_a + TF-IDF + PCA} & $0.0614\pm0.0025$ & $-0.0896\pm0.0350$ & \\
\hline
\end{tabular}
\caption{Results based on the 5-fold cross-validation procedure of the proposed methods for the response time and difficulty prediction tasks. To select the runs for each task, we employ the Kendall $\tau$ correlation. For each task, we highlight the top three Kendall $\tau$ correlations in {\color{red}\textbf{red (bold)}}, {\color{ForestGreen}green}, {\color{RoyalBlue}blue}, respectively. We highlight the best MSE for each task in bold. The $\downarrow$ and $\uparrow$ symbols indicate when lower or upper values are better, respectively. The $*$ symbol indicates the results are added post-competition.}
\label{tab_results}
\end{center}
\end{table*}

\subsection{Cross-Validation Results}

Due to the modest training dataset size (466 training samples), we employ 5-fold cross-validation to obtain robust evaluation results. We present the results based on the cross-validation procedure in Table~\ref{tab_results}. The results represent the average MSE and Kendall $\tau$ correlation values obtained across the 5 folds. Our experiments show a notable difference in task difficulty. Indeed, predicting difficulty proves to be significantly more challenging than predicting response time.

\begin{table*}[!t]
\setlength\tabcolsep{2.9pt}
\begin{center}
\begin{tabular}{|l|l|l|c|c|c|c|c|}
\hline
Task & Model & Features & RMSE $\downarrow$ & MSE $\downarrow$ & Kendall $\tau$ $\uparrow$ & Run & Rank \\
\hline 
\hline
\multirow{5}{*}{Response Time} & BERT & q\_answers  & $26.846$ &  $0.0333$ & $0.3579$ & 1 & 11/34 \\ 
\cline{2-8}
& BERT & q\_llms\_a & $26.768$ &  $0.0331$ & $0.3482$ & 2 & 10/34\\
\cline{2-8}
& GPT-2 & q\_answers & $26.073$ & $0.0366$ & $0.4767$ & 3 & 7/34\\
\cline{2-8}
& SVR & q\_llms\_a + BERT & $25.621$ & $0.0160$ & $0.4472$ & $*$& 5/35 \\
\cline{2-8}
& SVR & q\_llms\_a\_key + BERT & $25.613$ & $0.0160$ & $0.4399 $ & $*$& 5/35 \\
\hline
\hline
\multirow{5}{*}{Difficulty} & BERT & q\_llms\_a  & $0.308$ & $0.0654$ & $0.2179$ & 1 & 9/43\\ 
\cline{2-8}
& GPT-2 & q\_answers & $0.337$ & $0.1031$ & $0.0275$ & 2 & 34/43\\
\cline{2-8}
& GPT-2 & q\_llms\_a & $0.328$ & $0.1502$ & $0.0008$ & 3 & 30/43\\
\cline{2-8}
& SVR & q\_llms\_a + BERT & $0.292$  & $0.0638$ & $0.0517$ & $*$& 1/44 \\
\cline{2-8}
& SVR & q\_llms\_a\_key + BERT & $0.281$ & $0.0582$ & $0.1519$ & $*$ & 1/44\\
\hline
\end{tabular}
\caption{Test results of our best performing methods for the response time and difficulty prediction tasks. We report the official evaluation metric (RMSE), along with our metrics (MSE and Kendall $\tau$). The $\downarrow$ and $\uparrow$ symbols indicate when lower or upper values are better, respectively. The $*$ symbol indicates the results were added post-competition.}
\label{tab_test_mse_results}
\end{center}
\end{table*}

\noindent 
\textbf{Response time.} 
Our 5-fold cross-validation results indicate that the SVR+BERT models based on ``q\_llms\_a\_key'' (0.0132) and  ``q\_llms\_a'' (0.0134) features achieve the best MSE values for question response time prediction. They are followed by the fine-tuned BERT based on ``q\_answers'' features (0.0148). In terms of the Kendall $\tau$ correlation, the top three models are the same, but their ranking is different.
More precisely, the fine-tuned BERT based on ``q\_answers'' features surpasses the SVR+BERT models in terms of Kendall $\tau$.

We notice that the $\nu$-SVR models based on TF-IDF representations struggle to learn effective relationships between features and target labels. However, this is clearly an issue of the shallow TF-IDF features, since the $\nu$-SVR models based on BERT embeddings are at the opposite end of the performance spectrum.


Our experiments reveal two interesting findings regarding feature selection for response time prediction. First, transformer models that rely only on the multiple-choice answers obtain sub-optimal results compared with those that include the original question. This suggests that the question itself provides valuable information about the response time. The second important observation is that the AnswerKey feature becomes useful when combined with LLM answers, boosting the performance of SVR+BERT when using ``q\_llms\_a\_key'' features instead of `q\_llms\_a'' features, with respect to both MSE and Kendall $\tau$ measures.

\noindent
\textbf{Difficulty.} 
Similar to the response time prediction task, we analyze the top models for question difficulty prediction in terms of both MSE and Kendall $\tau$ correlation. Interestingly, the models achieving the best MSE scores, namely the fine-tuned BERT models based on ``q\_llms\_a'' (0.0500) and ``answers'' (0.0522) features, incorporate the correct answer information. However, in terms of Kendall $\tau$, the top models are slightly different. While the SVR+BERT with ``q\_llms\_a\_key'' features (0.1592) reaches the highest correlation, the second and third best models employ ``q\_llms\_a'' features in combination with BERT (0.1470) and GPT-2 (0.1259). Notably, all these models benefit from the inclusion of questions and LLM answers. Moreover, all but one of the top models for both metrics include the question text as input. This reinforces the importance of the question itself for predicting difficulty. Furthermore, the best Kendall $\tau$ scores are obtained by models that always incorporate both the question and LLM answers. This highlights the potential of LLMs in capturing nuances beyond the provided question and answer choices, leading to more accurate predictions. 

Similar to the previous task, the $\nu$-SVR models based on TF-IDF features seem to produce subpar results, given that their Kendall $\tau$ scores indicate negative correlations between predictions and target labels. However, the $\nu$-SVR models based on BERT embeddings achieve comparable results with the fine-tuned transformer-based approaches, and one of the former models (based on ``q\_llms\_a\_key'' features) performs even better in terms of Kendall $\tau$ than the top-three submitted models.

\subsection{Final Test Results}

For the test dataset, we report our two evaluation metrics, MSE and Kendall $\tau$, on the normalized labels, as well as the official evaluation metric, i.e.~the Root Mean Squared Error (RMSE), on the raw target labels. For the final evaluation on the official test set, we selected the top three models in terms of Kendall $\tau$ values. The corresponding results are presented in Table~\ref{tab_test_mse_results}. In the same table, we also include our post-competition results.

\noindent
\textbf{Response time.} All three submitted methods reach higher (worse) MSE values on the test set compared with the 5-fold cross-validation results, perhaps due to overfitting. The best MSE is achieved using the fine-tuned BERT model and the ``q\_llms\_a'' features (0.0331), surpassing the models based on ``q\_answers'' features. The MSE-based ranking of the three runs on the test set is not the same as the one obtained via cross-validation. The ranking based on the Kendall $\tau$ correlation is also different, with the best model on the test set being the fine-tuned GPT-2 based on ``q\_answers'' features (0.4767). This model also achieves better RMSE on the test set. However, for the other two submitted models, the RMSE metric is not correlated with Kendall $\tau$. Compared with the other competitors, our best model ranked 7th out of 34 models.

Our post-competition results obtained by the $\nu$-SVR+BERT models reveal consistent MSE and Kendall $\tau$ values across test and cross-validation evaluations. This suggests that keeping the pre-trained BERT frozen leads to a higher generalization capacity when the data available for fine-tuning is so small (less than 500 samples). Notably, calculating the RMSE on the held-out test set demonstrates that the SVR+BERT models outperform our officially submitted models, potentially obtaining a better rank (5th place out of 35).

\noindent
\textbf{Difficulty.} The MSE values of our final submissions for the difficulty prediction task are higher (worse) for two out of three methods, when compared with the values reported during the 5-fold cross-validation experiments. The respective methods are the fine-tuned GPT-2 based on ``q\_answers'' features and the fine-tuned GPT-2 based on ``q\_llms\_a'' features. The same two methods reach poor Kendall $\tau$ values, indicating almost no correlation between ground-truth and predicted labels. However, for our first run, which is represented by the fine-tuned BERT based on ``q\_llms\_a'' features, both MSE and Kendall $\tau$ values are comparable to the corresponding values reported using cross-validation (MSE: 0.0500 vs.~0.654, Kendall $\tau$: 0.1470 vs.~2179). Our findings are in line with the official results based on RMSE, which show that the fine-tuned BERT based on ``q\_llms\_a'' features is our best run. Compared with the models submitted by other participants, our best model for the question difficulty task ranks 9th out of 43 models.

Our post-competition $\nu$-SVR-based models yield superior performance compared to all three models submitted for the official evaluation. Remarkably, both post-competition models exhibit consistent MSE values between the cross-validation and test sets, hinting at the effective mitigation of overfitting which seems to affect our fine-tuned BERT and GPT-2 models. The configuration based on ``q\_llm\_a\_key'' features achieves the lowest MSE of 0.0582, followed closely by the configuration based on ``q\_llms\_a'' features, with an MSE of 0.0638. This further confirms the utility of the AnswerKey feature in combination with LLM answers. Furthermore, considering the official RMSE metric, our post-competition models achieve impressive results. The SVR+BERT based on ``q\_llm\_a\_key'' features attains the lowest RMSE of 0.281, followed by the version based on ``q\_llm\_a'' features with an RMSE of 0.292. These results would have positioned our post-competition models at the top of the leaderboard.




\section{Conclusion}

In this paper, we presented our approaches to the BEA 2024 Shared Task on Automated Prediction of Item Difficulty and Item Response Time of retired USMLE MCQs. Our main contribution is a task-specific data augmentation method based on adding answers to MCQs using LLMs prompted in a zero-shot setup. We carried out exhaustive experiments for both tasks, using two strong transformer-based models, in both fine-tuning and linear probing settings. We employed seven different types of feature combinations, while leveraging LLM-based answers. The empirical results showed four key findings. First, the difficulty prediction task is significantly harder than the response time prediction task. Second, we noticed that the top-performing approaches always made use of the question text. Third, LLM answers had a positive impact on performance, especially on the more difficult prediction task. Fourth, linear probing (training an SVR on frozen pre-trained features) shows a better generalization capacity than end-to-end fine-tuning, most likely due to the small training set available for the competition.

\section{Limitations}

To collect answers from the LLMs, we used a V100 GPT Colab runtime, with 78.2 GB Disk Space, which only allowed us to prompt the smallest versions of the three LLMs, each based on 7 billion parameters. Due to our resource limitations, we were not able to prompt larger LLMs, which could have led to better results. 

The limited number of samples was an important challenge for the evaluated transformers, which are prone to overfitting on small datasets. The final results indicate that our models suffered from some level of overfitting. In future work, we aim to study several ways to avoid overfitting, such as using dropout, frozen layers, regularization terms, etc.

\bibliography{custom}

\begin{thebibliography}{15}
\expandafter\ifx\csname natexlab\endcsname\relax\def\natexlab#1{#1}\fi

\bibitem[{Almazrouei et~al.(2023)Almazrouei, Alobeidli, Alshamsi, Cappelli,
  Cojocaru, Debbah, Goffinet, Hesslow, Launay, Malartic et~al.}]{falcon40b}
Ebtesam Almazrouei, Hamza Alobeidli, Abdulaziz Alshamsi, Alessandro Cappelli,
  Ruxandra Cojocaru, M{\'e}rouane Debbah, {\'E}tienne Goffinet, Daniel Hesslow,
  Julien Launay, Quentin Malartic, et~al. 2023.
\newblock \href {https://arxiv.org/pdf/2311.16867.pdf} {{The Falcon series of
  open language models}}.
\newblock \emph{arXiv preprint arXiv:2311.16867}.

\bibitem[{Baldwin et~al.(2021)Baldwin, Yaneva, Mee, Clauser, and
  Ha}]{Baldwin-JEM-2021}
Peter Baldwin, Victoria Yaneva, Janet Mee, Brian~E. Clauser, and Le~An Ha.
  2021.
\newblock \href {https://doi.org/https://doi.org/10.1111/jedm.12264} {Using
  natural language processing to predict item response times and improve test
  construction}.
\newblock \emph{Journal of Educational Measurement}, 58(1):4--30.

\bibitem[{Chen et~al.(2023)Chen, Hernández-Cano, Romanou, Bonnet, Matoba,
  Salvi, Pagliardini, Fan, Köpf, Mohtashami, Sallinen, Sakhaeirad, Swamy,
  Krawczuk, Bayazit, Marmet, Montariol, Hartley, Jaggi, and
  Bosselut}]{chen2023meditron70b}
Zeming Chen, Alejandro Hernández-Cano, Angelika Romanou, Antoine Bonnet, Kyle
  Matoba, Francesco Salvi, Matteo Pagliardini, Simin Fan, Andreas Köpf,
  Amirkeivan Mohtashami, Alexandre Sallinen, Alireza Sakhaeirad, Vinitra Swamy,
  Igor Krawczuk, Deniz Bayazit, Axel Marmet, Syrielle Montariol, Mary-Anne
  Hartley, Martin Jaggi, and Antoine Bosselut. 2023.
\newblock \href {https://arxiv.org/pdf/2311.16079.pdf} {{MEDITRON-70B: Scaling
  Medical Pretraining for Large Language Models}}.
\newblock \emph{arXiv preprint arXiv:2311.16079}.

\bibitem[{Devlin et~al.(2019)Devlin, Chang, Lee, and
  Toutanova}]{DBLP:journals/corr/abs-1810-04805}
Jacob Devlin, Ming-Wei Chang, Kenton Lee, and Kristina Toutanova. 2019.
\newblock \href {https://www.aclweb.org/anthology/N19-1423} {{BERT:
  Pre-training of Deep Bidirectional Transformers for Language Understanding}}.
\newblock In \emph{Proceedings of the 2019 Conference of the North American
  Chapter of the Association for Computational Linguistics: Human Language
  Technologies}, pages 4171--4186.

\bibitem[{Georgescu et~al.(2023)Georgescu, Ionescu, and
  Miron}]{georgescu2023diversity}
Mariana-Iuliana Georgescu, Radu~Tudor Ionescu, and Andreea~Iuliana Miron. 2023.
\newblock \href {https://doi.org/10.1145/3555776.3577682} {Diversity-promoting
  ensemble for medical image segmentation}.
\newblock In \emph{Proceedings of the 38th ACM/SIGAPP Symposium on Applied
  Computing}, pages 599--606, New York, NY, USA. Association for Computing
  Machinery.

\bibitem[{Ha et~al.(2019)Ha, Yaneva, Baldwin, and
  Mee}]{ha-etal-2019-predicting}
Le~An Ha, Victoria Yaneva, Peter Baldwin, and Janet Mee. 2019.
\newblock \href {https://doi.org/10.18653/v1/W19-4402} {Predicting the
  difficulty of multiple choice questions in a high-stakes medical exam}.
\newblock In \emph{Proceedings of the Fourteenth Workshop on Innovative Use of
  NLP for Building Educational Applications}, pages 11--20, Florence, Italy.
  Association for Computational Linguistics.

\bibitem[{Jiang et~al.(2023)Jiang, Sablayrolles, Mensch, Bamford, Chaplot,
  de~las Casas, Bressand, Lengyel, Lample, Saulnier, Lavaud, Lachaux, Stock,
  Scao, Lavril, Wang, Lacroix, and Sayed}]{jiang2023mistral}
Albert~Q. Jiang, Alexandre Sablayrolles, Arthur Mensch, Chris Bamford,
  Devendra~Singh Chaplot, Diego de~las Casas, Florian Bressand, Gianna Lengyel,
  Guillaume Lample, Lucile Saulnier, Lélio~Renard Lavaud, Marie-Anne Lachaux,
  Pierre Stock, Teven~Le Scao, Thibaut Lavril, Thomas Wang, Timothée Lacroix,
  and William~El Sayed. 2023.
\newblock \href {https://arxiv.org/pdf/2310.06825.pdf} {{Mistral 7B}}.
\newblock \emph{arXiv preprint arXiv:2310.06825}.

\bibitem[{Loshchilov and Hutter(2019)}]{Loshchilov-ICLR-2019}
Ilya Loshchilov and Frank Hutter. 2019.
\newblock \href {https://openreview.net/pdf?id=Bkg6RiCqY7} {{Decoupled Weight
  Decay Regularization}}.
\newblock In \emph{Proceedings of the International Conference on Learning
  Representations}.

\bibitem[{Radford et~al.(2019)Radford, Wu, Child, Luan, Amodei, and
  Sutskever}]{radford2019language}
Alec Radford, Jeff Wu, Rewon Child, David Luan, Dario Amodei, and Ilya
  Sutskever. 2019.
\newblock \href
  {https://cdn.openai.com/better-language-models/language_models_are_unsupervised_multitask_learners.pdf}
  {Language models are unsupervised multitask learners}.
\newblock \emph{OpenAI blog}, 1(8):9.

\bibitem[{Sch{\"o}lkopf et~al.(2000)Sch{\"o}lkopf, Smola, Williamson, and
  Bartlett}]{Scholkopf-NC-2000}
Bernhard Sch{\"o}lkopf, Alex~J. Smola, Robert~C. Williamson, and Peter~L.
  Bartlett. 2000.
\newblock \href {https://doi.org/https://doi.org/10.1162/089976600300015565}
  {New support vector algorithms}.
\newblock \emph{Neural computation}, 12(5):1207--1245.

\bibitem[{Settles et~al.(2020)Settles, LaFlair, and
  Hagiwara}]{settles-etal-2020-machine}
Burr Settles, Geoffrey~T. LaFlair, and Masato Hagiwara. 2020.
\newblock \href {https://doi.org/10.1162/tacl_a_00310} {Machine
  learning{--}driven language assessment}.
\newblock \emph{Transactions of the Association for Computational Linguistics},
  8:247--263.

\bibitem[{Xue et~al.(2020)Xue, Yaneva, Runyon, and
  Baldwin}]{xue-etal-2020-predicting}
Kang Xue, Victoria Yaneva, Christopher Runyon, and Peter Baldwin. 2020.
\newblock \href {https://doi.org/10.18653/v1/2020.bea-1.20} {Predicting the
  difficulty and response time of multiple choice questions using transfer
  learning}.
\newblock In \emph{Proceedings of the Fifteenth Workshop on Innovative Use of
  NLP for Building Educational Applications}, pages 193--197, Seattle, WA, USA
  → Online. Association for Computational Linguistics.

\bibitem[{Yaneva et~al.(2020)Yaneva, Ha, Baldwin, and
  Mee}]{yaneva-etal-2020-predicting}
Victoria Yaneva, Le~An Ha, Peter Baldwin, and Janet Mee. 2020.
\newblock \href {https://aclanthology.org/2020.lrec-1.841} {Predicting item
  survival for multiple choice questions in a high-stakes medical exam}.
\newblock In \emph{Proceedings of the Twelfth Language Resources and Evaluation
  Conference}, pages 6812--6818, Marseille, France. European Language Resources
  Association.

\bibitem[{Yaneva et~al.(2021)Yaneva, Jurich, Ha, and
  Baldwin}]{yaneva-etal-2021-using}
Victoria Yaneva, Daniel Jurich, Le~An Ha, and Peter Baldwin. 2021.
\newblock \href {https://aclanthology.org/2021.bea-1.23} {Using linguistic
  features to predict the response process complexity associated with answering
  clinical {MCQ}s}.
\newblock In \emph{Proceedings of the Fifteenth Workshop on Innovative Use of
  NLP for Building Educational Applications}, pages 223--232, Online.
  Association for Computational Linguistics.

\bibitem[{Yaneva et~al.(2024)Yaneva, North, Baldwin, Le, Rezayi, Zhou,
  Choudhury, Harik, and Clauser}]{yaneva-etal-2024-DART-MCQ}
Victoria Yaneva, Kai North, Peter Baldwin, An~Ha Le, Saed Rezayi, Yiyun Zhou,
  Sagnik~Ray Choudhury, Polina Harik, and Brian Clauser. 2024.
\newblock {{Findings from the First Shared Task on Automated Prediction of
  Difficulty and Response Time for Multiple Choice Questions}}.
\newblock In \emph{Proceedings of the 19th Workshop on Innovative Use of NLP
  for Building Educational Applications (BEA 2024)}, Mexico City, Mexico.
  Association for Computational Linguistics.

\end{thebibliography}

\end{document}